\documentclass{article}
\usepackage{spconf,amsmath,graphicx}
\usepackage{multirow}
\usepackage{float}
\usepackage{stfloats}
\usepackage{graphicx}
\usepackage{afterpage}
\usepackage{graphicx}
\usepackage{capt-of}
\usepackage{enumitem}
\usepackage[numbers]{natbib}
\usepackage{booktabs}
\usepackage{graphicx}
\usepackage{caption}
\usepackage{subcaption}
\usepackage{amssymb}
\usepackage{enumitem}
\usepackage[hidelinks]{hyperref} %
\setlist[itemize]{itemsep=0pt, parsep=0pt, topsep=2pt, partopsep=0pt, leftmargin=1.2em}

\newcommand{\best}[1]{\textbf{#1}}
\newcommand{\second}[1]{\underline{#1}}

\usepackage{afterpage}

\title{CLIP-Guided Unsupervised Semantic-Aware Exposure Correction}
\name{\begin{tabular}{c}
Puzhen Wu$^{1,2}$ \qquad Han Weng$^{2}$ \qquad Quan Zheng$^{1}$ \qquad Yi Zhan$^{3}$ \qquad
Hewei Wang$^{2}$ \qquad Yiming Li$^{2}$ \\
Jiahui Han$^{2}$ \qquad Rui Xu$^{3}$
\end{tabular}}

\address{\begin{tabular}{c}
$^{1}$Institute of Software, Chinese Academy of Sciences \\
$^{2}$Beijing-Dublin Internationl College, University College Dublin \\
$^{3}$School of Computer Science, Peking University \\
{\tt puzhenwu8@connect.hku.hk}
\end{tabular}}

\begin{document}
\ninept
\maketitle
\begin{abstract}
Improper exposure often leads to severe loss of details, color distortion, and reduced contrast. Exposure correction still faces two critical challenges: (1) the ignorance of object-wise regional semantic information causes the color shift artifacts; (2) real-world exposure images generally have no ground-truth labels, and its labeling entails massive manual editing. To tackle the challenges, we propose a new unsupervised semantic-aware exposure correction network. It contains an adaptive semantic-aware fusion module, which effectively fuses the semantic information extracted from a pre-trained Fast Segment Anything Model into a shared image feature space. Then the fused features are used by our multi-scale residual spatial mamba group to restore the details and adjust the exposure. To avoid manual editing, we propose a pseudo-ground truth generator guided by CLIP, which is fine-tuned to automatically identify exposure situations and instruct the tailored corrections. Also, we leverage the rich priors from the FastSAM and CLIP to develop a semantic-prompt consistency loss to enforce semantic consistency and image-prompt alignment for unsupervised training. Comprehensive experimental results illustrate the effectiveness of our method in correcting real-world exposure images and outperforms state-of-the-art unsupervised methods both numerically and visually.
\end{abstract}

\vspace{-8pt}
\section{Introduction}
\label{sec:intro}

Real-world imaging is often constrained by diverse environmental lighting and improper camera exposure settings (e.g., aperture, shutter speed, and ISO), leading to exposure issues in images. These problems result in diminished detail, color distortion, and reduced contrast, significantly degrading both image quality and visual appeal. Manual exposure adjustment, however, is time-consuming and labor-intensive, highlighting the need for an automated exposure correction method that can effectively enhance the visibility of image details and improve overall image quality.

Recently, many deep learning methods~\cite{31,45,46} have been proposed for exposure correction, yet two challenges persist. \textbf{(1)} Most approaches apply global adjustments~\cite{31,46,45} or coarse binary masks~\cite{45}, ignoring region-level semantics; this breaks semantic consistency, degrades structural fidelity, and induces color shifts. \textbf{(2)} Many rely on supervised learning~\cite{45,46,31}, requiring expert-edited targets that are costly and subjective, yielding small labeled datasets and poor generalization to real-world unlabeled images.

\begin{figure}[t!]
    \centering
    \includegraphics[width=\linewidth]{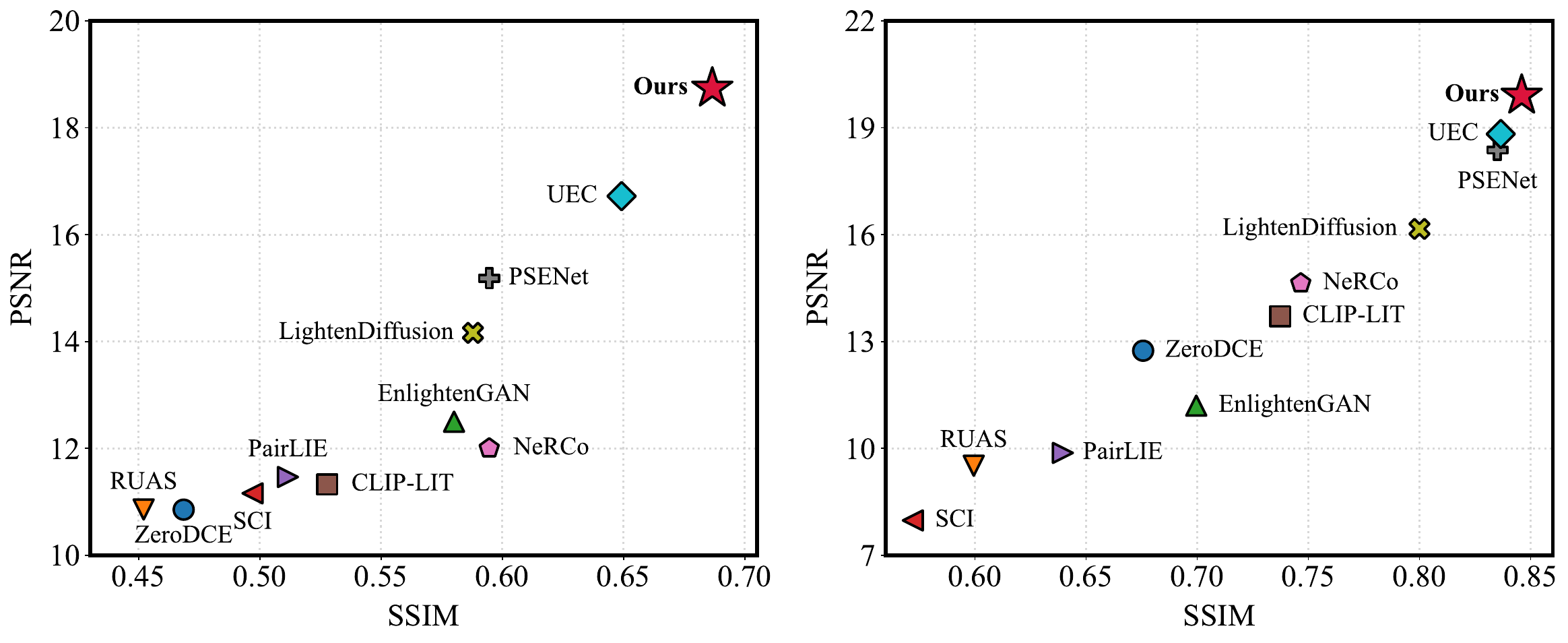}
    \vspace{-18pt}
    \caption{Performance comparison with other unsupervised methods.}
    \vspace{-18pt}
    \label{fig:comparison1}
\end{figure}

For the first challenge, We leverage semantic features extracted from the Fast Segment Anything Model (FastSAM) \cite{fastsam} to effectively introduce object-level semantic priors, by which object-wise regional exposure variations can be well captured. 
Further, we introduce a new semantic-aware exposure correction network. It consists of multi-scale Semantics-Informed Mamba Reconstruction (SIMR) blocks with downsampling and upsampling operations. Within each SIMR block, we design an Adaptive Semantic-Aware Fusion (ASF) module to fuse the semantic features with image-space features. Following the ASF module, we devise a Residual Spatial Mamba Group (RSMG) module to refine exposure correction with capturing long-range spatial dependencies. To tackle the second challenge, we introduce a CLIP-guided pseudo-ground truth (GT) generator. It employs fine-tuned CLIP \cite{radford2021learning} prompts to automatically classify exposed images into different exposure situations, which then guide the tailored gamma corrections to form pseudo well-exposed GT. By leveraging pseudo-GT, we eliminate laborious manual labeling. Additionally, we design a new semantic-prompt consistency loss to promote semantic consistency in the semantic space, and push the results of our unsupervised correction towards the well-exposed images in the vision-language joint space.
Extensive experiments verify that our method outperforms SOTA unsupervised methods in quantitative metrics (Fig. \ref{fig:comparison1}), highlighting its effectiveness for exposure correction. Our main contributions are threefold:

\begin{itemize}[itemsep=0pt, parsep=0pt, topsep=2pt, partopsep=0pt, leftmargin=1.2em]
    \item We propose a novel semantic-aware exposure correction network, which achieves semantics-informed reconstruction with SIMR modules by effectively introducing object-level semantic priors.
    \item We design an automated pseudo-GT generation approach, eliminating the need for laborious manual labeling.
    \item We propose a loss that enforces semantic consistency and image–prompt alignment for high-quality unsupervised training.
\end{itemize}

\begin{figure*}[htp]
    \centering
    \includegraphics[width=1.0\linewidth]{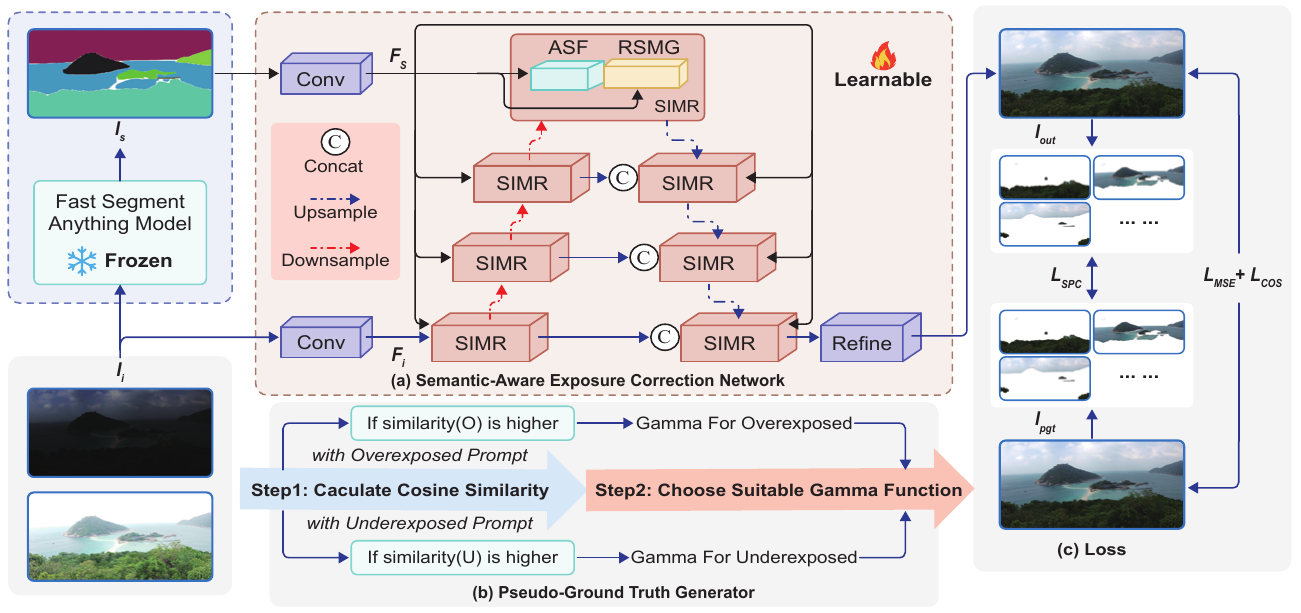}
    \vspace{-15pt}
    \caption{Overview of our framework: (a) The semantic-aware exposure correction network. (b) Pseudo-ground truth generator. (c) Loss.}
    \label{fig:net}
    \vspace{-12pt}
\end{figure*}

\vspace{-10pt}
\section{Related Work}

\label{sec:format}

\noindent\textbf{Traditional Methods.}
Early approaches enhance contrast via histogram equalization and gamma correction, or use Retinex-based decomposition of illumination/reflectance~\cite{16}. While effective in specific cases, these hand-crafted priors often yield artifacts and lack robustness across diverse scenes.

\noindent \textbf{Supervised Learning-based Methods.}
Supervised methods tackle under/over exposure~\cite{45,46,31} or focus on low-light~\cite{bib9,bib14,bib16,bib17,20}, but can suffer content loss and color shifts~\cite{bib9}. Representative designs include Laplacian pyramids~\cite{29}, exposure-consistency modules~\cite{36}, generative priors, exposure-invariant spaces, and spatial–frequency interactions~\cite{31}. Recent low-light enhancement works also emphasize efficient and lightweight designs for practical deployment~\cite{wang2025lightllie}, and source-free scene adaptation has been explored to improve robustness across real-world conditions without target labels~\cite{wang2025revitalize}. However, many still rely on global correction or coarse masks, struggling in mixed-lighting scenes~\cite{46}. Wu et al.~\cite{43} use semantics but only enhance underexposed regions, while RECNet~\cite{45} depends on binary masks.

\noindent \textbf{Unsupervised Learning-based Methods.}
To avoid costly labels, unsupervised methods~\cite{bib10,bib9,bib22,lightdiff} mainly target low-light and struggle with overexposure. Related label-free paradigms also include zero-shot adaptive enhancement via Retinex decomposition and curve estimation~\cite{xia2023zero}. PSENet~\cite{42} addresses mixed exposures but uses a fixed global parameter, risking artifacts; UEC~\cite{47} leverages multi-exposure sequences. These approaches largely ignore region-wise semantics, leading to local color shifts. We instead propose an unsupervised framework that integrates semantic cues with vision–language priors to handle mixed lighting.

\vspace{-10pt}
\section{Method}
\label{sec:pagestyle}

\vspace{-4pt}
\subsection{Semantic-Aware Exposure Correction 
Network}\label{sec:network}
\begin{figure}[t]
    \centering

    \begin{subfigure}[b]{0.3\linewidth}
        \includegraphics[width=\linewidth]{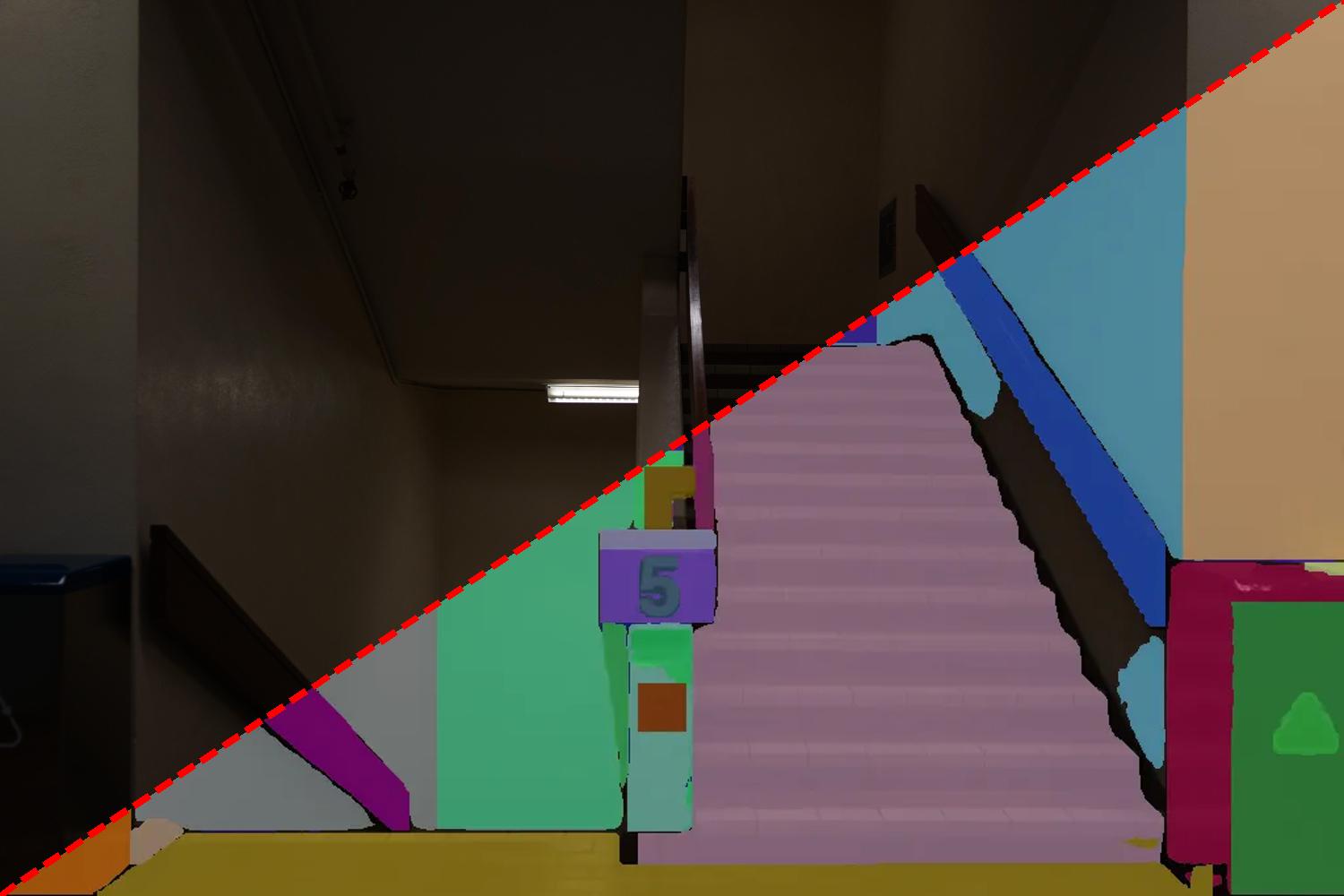} \\
        \vspace{-10pt}
        \caption{Underexposed}
    \end{subfigure}
    \begin{subfigure}[b]{0.3\linewidth}
        \includegraphics[width=\linewidth]{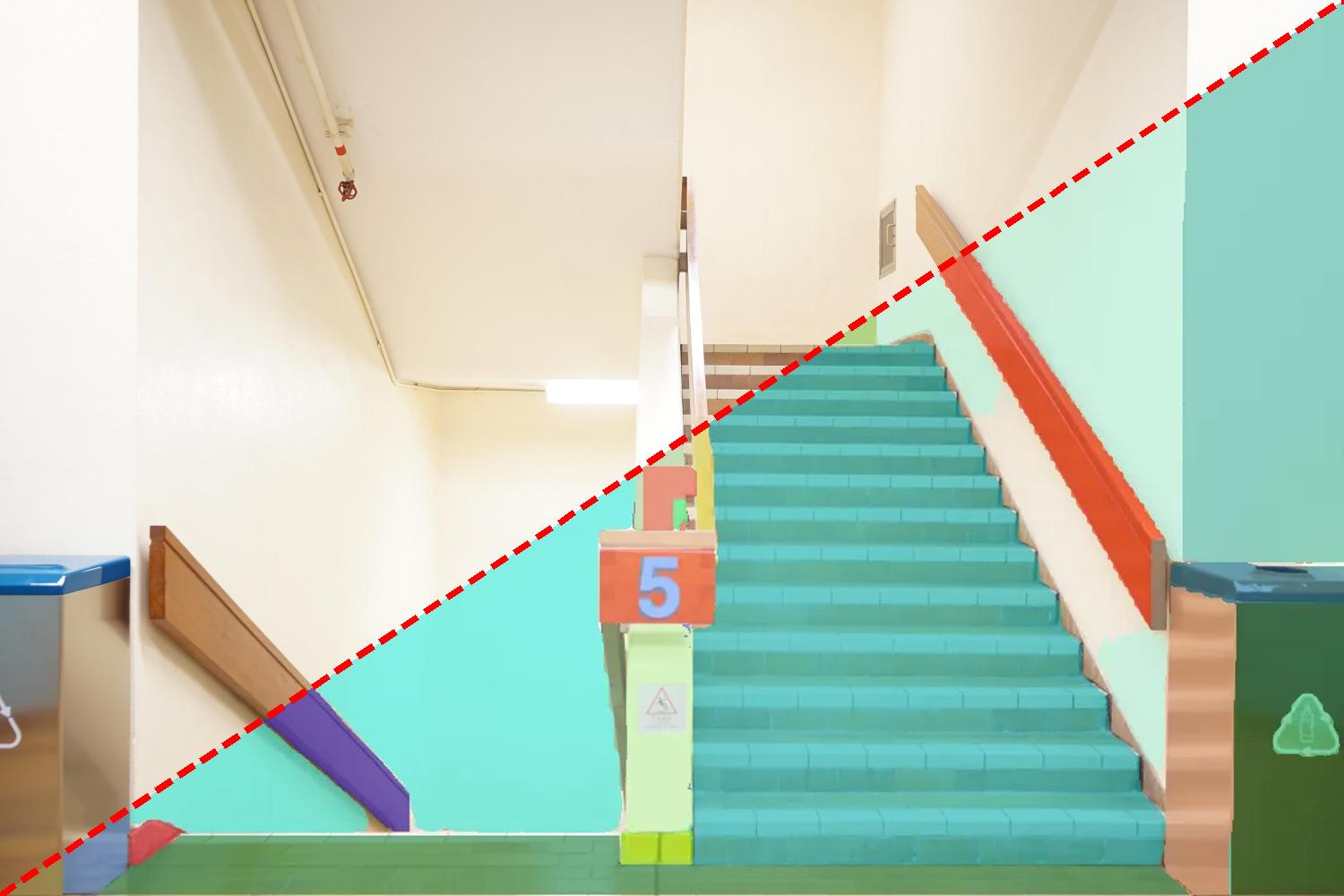} \\
        \vspace{-10pt}
        \caption{Overexposed}
    \end{subfigure}
    \begin{subfigure}[b]{0.3\linewidth}
        \includegraphics[width=\linewidth]{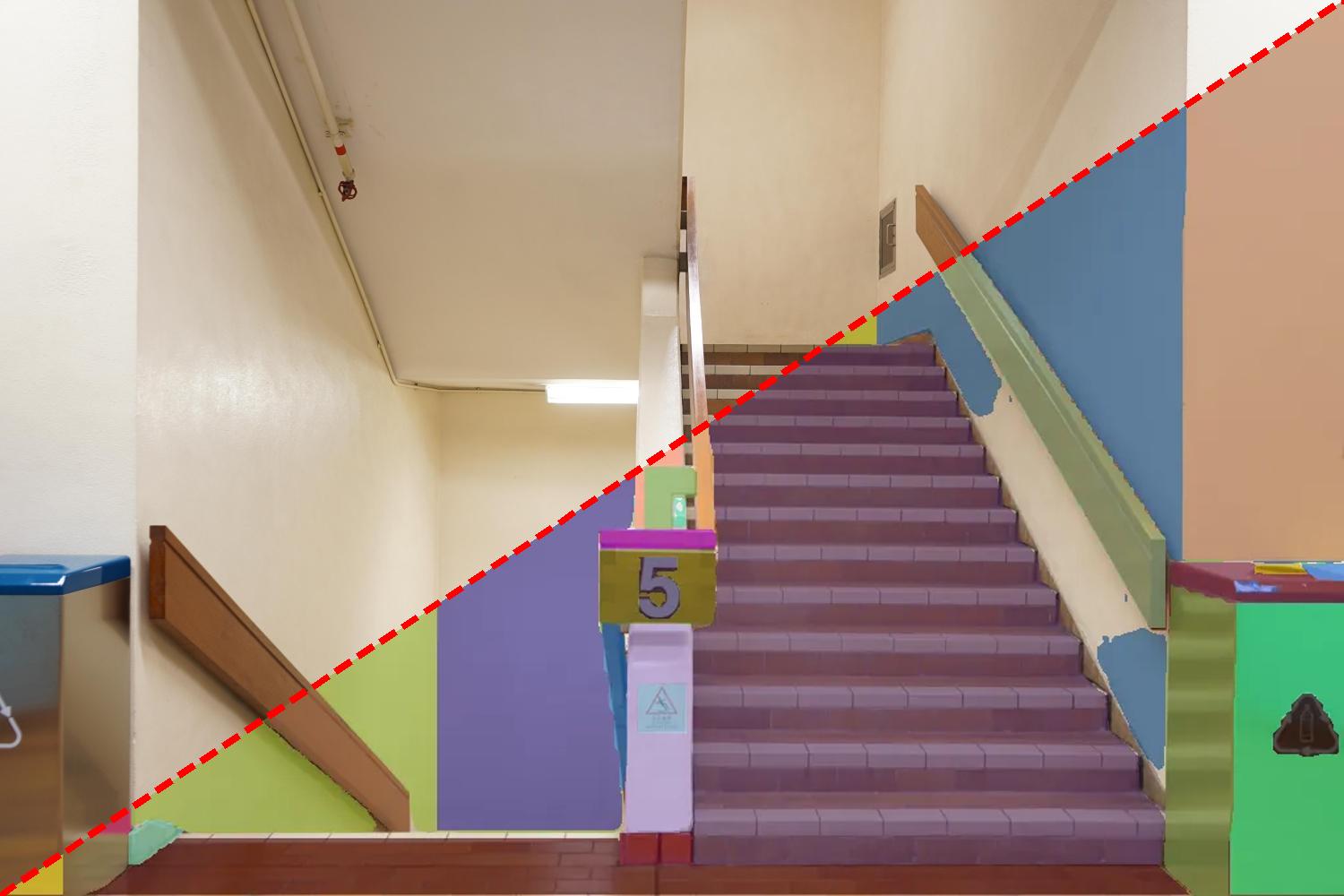} \\
        \vspace{-10pt}
        \caption{Well-exposed}
    \end{subfigure}
    \vspace{-8pt}
    \caption{FastSAM \cite{fastsam} segmentation results.}
    \vspace{-20pt}
    \label{fig:fastsam}
\end{figure}

\begin{figure}[h]
    \centering
    \includegraphics[width=1.0\linewidth]{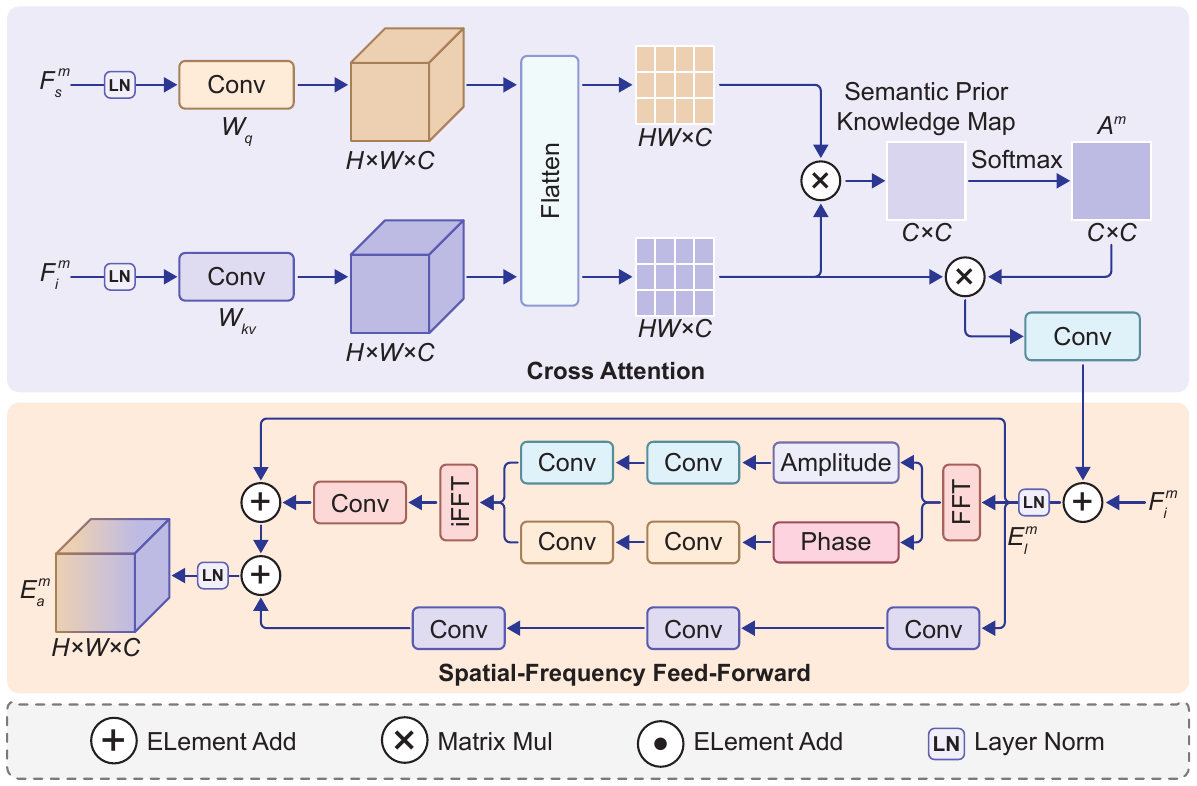}
    \caption{
    The architecture of the ASF module.
    }
    \vspace{-12pt}
    \label{fig:spk}
\end{figure}

\begin{figure}[h]
    \centering
    \includegraphics[width=1.0\linewidth]{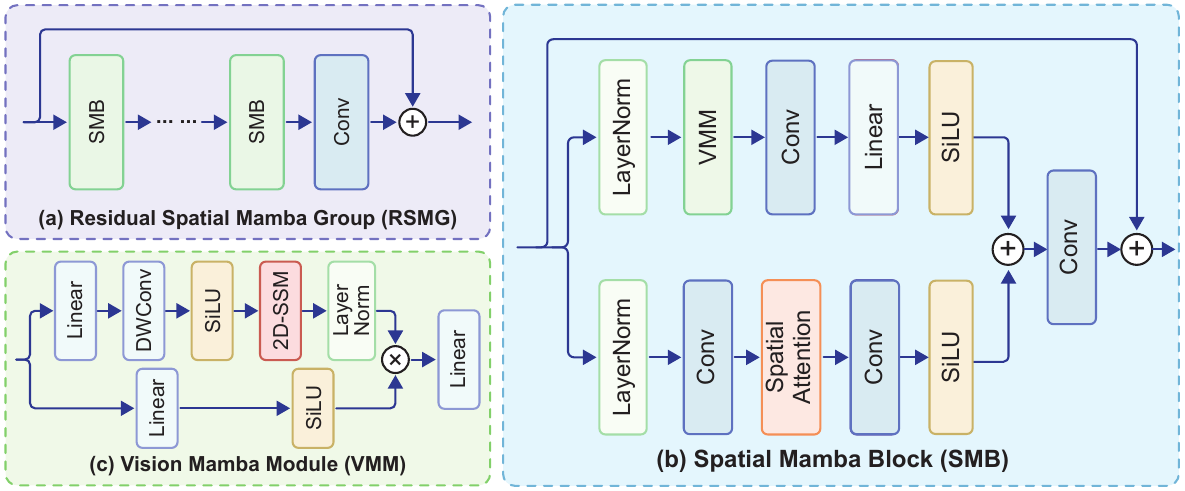}
    \vspace{-15pt}
    \caption{(a) Residual Spatial Mamba Group (RSMG), (b) Spatial Mamba Block (SMB), (c) Vision Mamba Module (VMM).}
    \vspace{-10pt}
    \label{fig:MambaIR}
\end{figure}

\vspace{-4pt}
Our network stacks SIMR blocks (Fig.~\ref{fig:net}) to perform multi-scale correction via down/up-sampling. Each SIMR contains an Adaptive Semantic-Aware Fusion (ASF, Fig.~\ref{fig:spk}) and a Residual Spatial Mamba Group (RSMG, Fig.~\ref{fig:MambaIR}). ASF injects region-level semantics to model exposure variations, while RSMG applies adaptive local correction to preserve spatial coherence. We obtain semantics with FastSAM (Fig. \ref{fig:fastsam}): for an image \(I_i\!\in\!\mathbb{R}^{H\times W\times 3}\), a segmentation map \(I_s=\mathbf{F}_{\text{sam}}(I_i)\) (frozen weights) is computed, then passed through a convolutional layer to produce semantic features \(F_s\). Downsampled \(F_s\) is fed to each SIMR scale (Fig.~\ref{fig:net}), enabling content-aware exposure adjustment under mixed lighting.

\vspace{0.5em}
\noindent\textbf{Adaptive Semantic-Aware Fusion Module.} As shown in Figure \ref{fig:spk}, our ASF module takes in the \( F_{i}^{m} \) and \( F_{s}^{m} \), which are the image-space feature and semantic feature for the $m$-th SIMR scale, respectively. Drawing inspiration from Wu et al.~\cite{43}, we first employ a cross-attention block for projecting the semantic feature and the image-space feature into a shared feature space. Specifically, the weighted feature map \( W_{kv}(LN(F_i^m)) \) is element-wise multiplied by the attention map \( A^m \) and goes through a convolutional layer to get the intermediate feature $F_{med}$, where \( W_{kv} \) is a convolutional layer and \(LN \) is layer normalization. $F_{med}$ and the skip-connected \( F_{i}^{m} \) are element-wise added and layer-normalized to obtain $E_l^m$.

To better fuse semantic and image features, we use a spatial–frequency feed-forward block. 
The \emph{frequency} branch applies FFT to \(E_l^m\), separates amplitude/phase, refines each with two \(1{\times}1\) convs, then uses Inverse FFT (IFFT) and a conv, with a residual from \(E_l^m\), to yield \(E_o^m\) and capture global exposure trends. 
The \emph{spatial} branch sends \(E_l^m\) through three convolutions to model local details. 
The two outputs are summed and normalized:

\begin{equation}
E_a^m = \mathrm{LN}\!\Big( E_l^m + \mathrm{FP}(E_l^m) + \mathrm{SP}(E_l^m) \Big),
\end{equation}
where \(\mathrm{FP}\) and \(\mathrm{SP}\) denote the frequency and spatial branches, respectively; \(E_a^m\) is then fed to the RSMG.

\begin{figure*}[t]
    \centering
    \begin{subfigure}[b]{0.137\textwidth}
        \includegraphics[width=\linewidth]{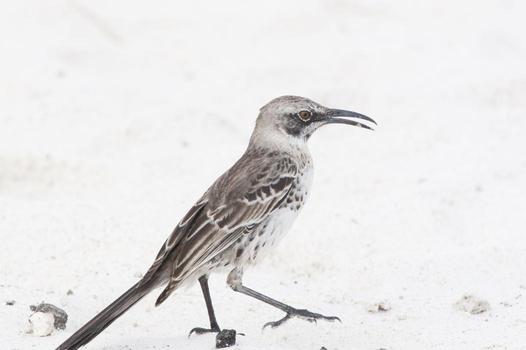}\\[1pt]
        \includegraphics[width=\linewidth]{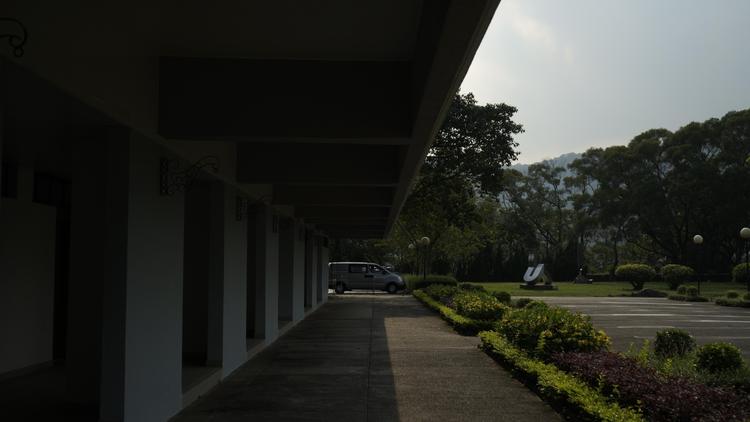}
        \caption{Input}
    \end{subfigure}
    \begin{subfigure}[b]{0.137\textwidth}
        \includegraphics[width=\linewidth]{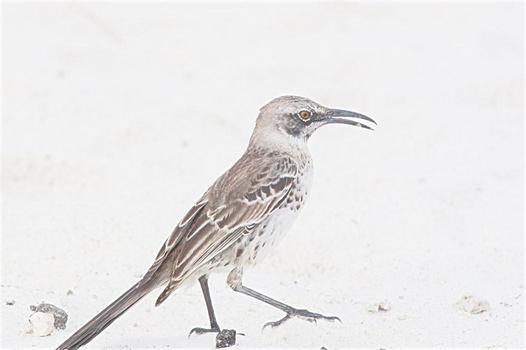}\\[1pt]
        \includegraphics[width=\linewidth]{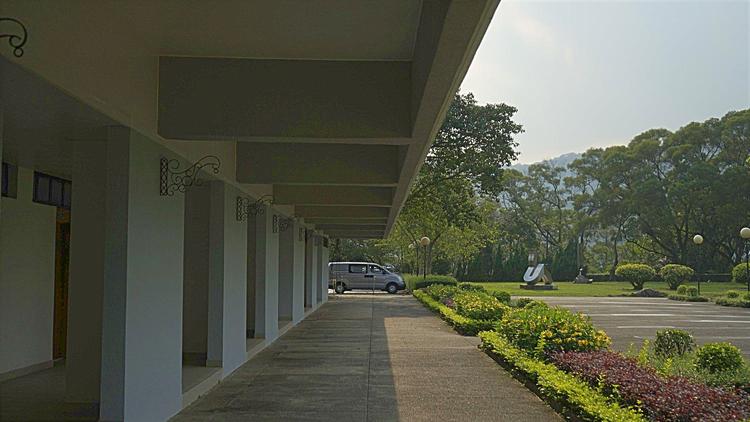}
        \caption{CLIP-LIT \cite{bib22}}
    \end{subfigure}
    \begin{subfigure}[b]{0.137\textwidth}
        \includegraphics[width=\linewidth]{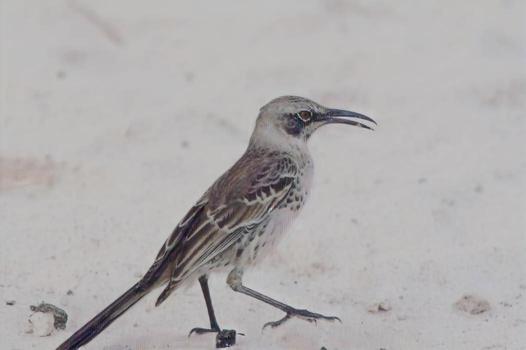}\\[1pt]
        \includegraphics[width=\linewidth]{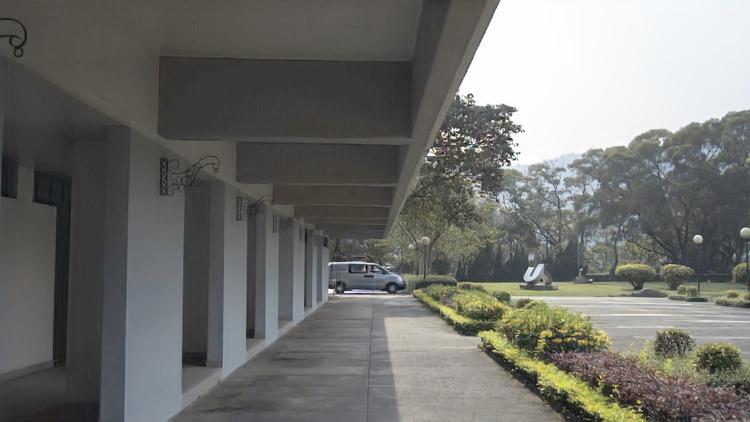}
        \caption{NeRCo \cite{20}}
    \end{subfigure}
    \begin{subfigure}[b]{0.137\textwidth}
        \includegraphics[width=\linewidth]{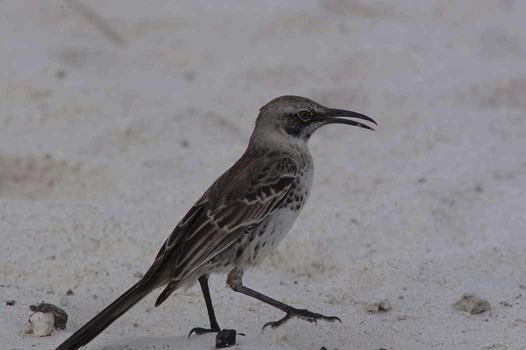}\\[1pt]
        \includegraphics[width=\linewidth]{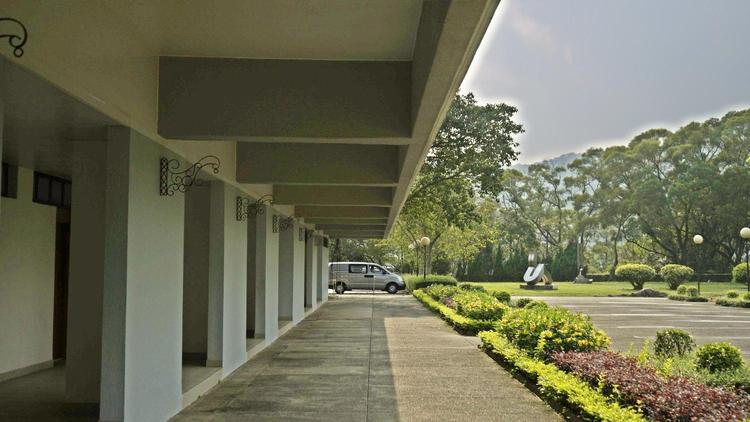}
        \caption{PSENet \cite{42}}
    \end{subfigure}
    \begin{subfigure}[b]{0.137\textwidth}
        \includegraphics[width=\linewidth]{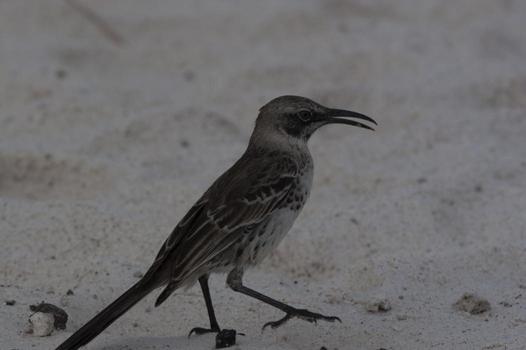}\\[1pt]
        \includegraphics[width=\linewidth]{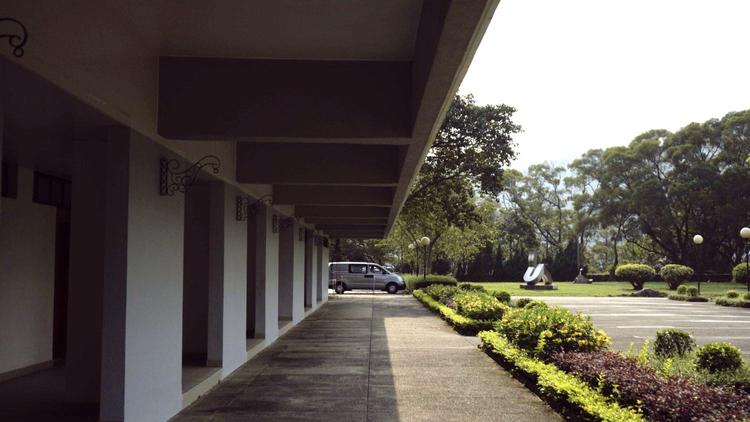}
        \caption{UEC \cite{47}}
    \end{subfigure}
    \begin{subfigure}[b]{0.137\textwidth}
        \includegraphics[width=\linewidth]{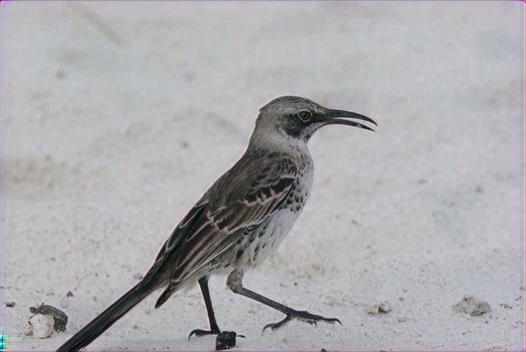}\\[1pt]
        \includegraphics[width=\linewidth]{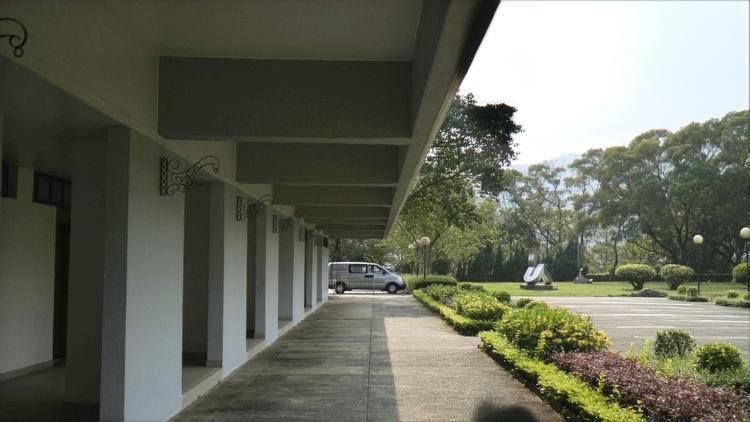}
        \caption{\textbf{Ours}}
    \end{subfigure}
    \begin{subfigure}[b]{0.137\textwidth}
        \includegraphics[width=\linewidth]{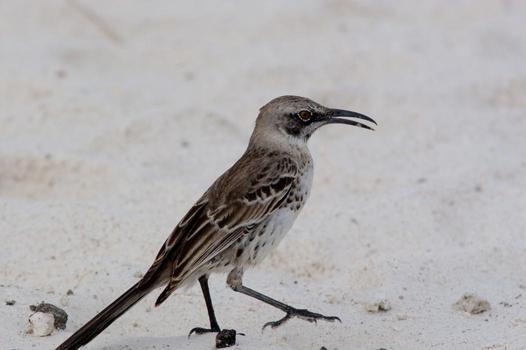}\\[1pt]
        \includegraphics[width=\linewidth]{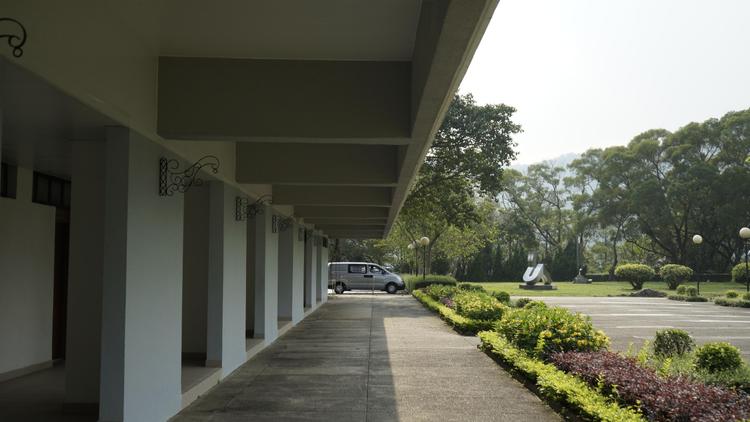}
        \caption{Reference}
    \end{subfigure}
    \vspace{-5pt}

    \caption{Qualitative comparison with state-of-the-art unsupervised methods.}
    \label{fig:comparison}
\end{figure*}

\vspace{0.5em}
\noindent\textbf{Residual Spatial Mamba Group.} We follow VMamba~\cite{liu2024vmamba} and remove the S6 in Mamba~\cite{mamba} and incorporate the 2D selective-scan module (2D-SSM), yielding a Vision Mamba Module (VMM; Fig.~\ref{fig:MambaIR}(c)). Mamba-style architectures have been explored in multiple areas~\cite{wu2025disease,wu2025amd}. 2D-SS unfolds a feature map along multiple directions into 1D sequences processed by state-space transitions, then folds them back to 2D, achieving a global receptive field with linear complexity. For exposure correction, this multi-directional modeling captures long-range spatial dependencies. On top of VMM, we propose a Spatial Mamba Block (SMB; Fig.~\ref{fig:MambaIR}(b)) that augments VMM with spatial attention to better handle spatially varying exposures.

We stack multiple Spatial Mamba Blocks (SMBs) into a residual group (RSMG; Fig.~\ref{fig:MambaIR}(a)). For the $l$-th SMB with input $F^{l}$, the upper branch normalizes $F^{l}$, applies VMM, a conv, and a linear layer to yield $D^{l}_{1}$; the lower branch applies normalization, convolutions, spatial attention, and convolutions to produce $D^{l}_{2}$. We fuse them by element-wise addition and add a skip from $F^{l}$ to obtain $F^{l+1}=F^{l}+D^{l}_{1}+D^{l}_{2}$. Stacking SMBs enhances exposure consistency and suppresses color shifts across regions.

\vspace{-8pt}
\subsection{CLIP Guided Pseudo-GT Generation}\label{sec:generator}
\vspace{-4pt}

\vspace{0.5em}
\noindent\textbf{Prompt Fine-tuning.} Inspired by CLIP-LIT \cite{bib22}, we start by initializing three prompts $T_w$, $T_u$, and $T_o$, corresponding to the well-exposed, underexposed, and overexposed conditions. Compared to only learning two prompts (for normal and backlit images) \cite{bib22}, our method extends to three dedicated prompts to better capture the diverse illumination distributions in real scenes. We then fine-tune these prompts using unpaired images from our dataset (see Section \ref{sec:Experimental Setup}), categorizing them into well-exposed ($I_w$), underexposed ($I_u$), and overexposed ($I_o$) groups. Each image is processed by the pre-trained CLIP image encoder $\Phi_{\text{image}}$ to produce latent vectors, while the prompts $T_w$, $T_u$, and $T_o$ are encoded by the CLIP text encoder $\Phi_{\text{text}}$ to obtain their latent vectors. Fine-tuning is guided by image-prompt similarity in the shared CLIP latent space, where a triplet cross-entropy loss function $\mathcal{L}_{\mathrm{tune}}$ encourages each prompt to closely align with its matching image category while minimizing similarity with unrelated categories, expressed as:

\vspace{-5pt}
\begin{equation}
\mathcal{L}_{\mathrm{tune}} = - \sum_{i \in \{w, u, o\}} y_i \cdot \log(\hat{y}_i) ,
\end{equation}
\vspace{-5pt}
\begin{equation}
\hat{y}_i 
= \frac{\exp\!\Bigl(\cos\bigl(\Phi_{\text{image}}(I),\,\Phi_{\text{text}}\bigl(T_i\bigr)\bigr)\Bigr)}
       {\sum_{j \in \{w, u, o\}} \exp\!\Bigl(\cos\bigl(\Phi_{\text{image}}(I),\,\Phi_{\text{text}}\bigl(T_j\bigr)\bigr)\Bigr)}.
\end{equation}
\vspace{-5pt}
\noindent where \( I \in \{ I_w, I_u, I_o \} \), \( y \) is the label of the current image.

\vspace{0.5em}
\noindent\textbf{Pseudo-GT Generator.} After fine-tuning the underexposed prompts $T_u$ and overexposed prompts $T_o$, we can calculate the similarity between an input image $I$ and each of these prompts as following: 
\vspace{-5pt}
\begin{equation}
sim_i = \cos(\Phi_{\text{image}}(I), \Phi_{\text{text}}(T_i)) ,
\end{equation}
where \( i \in \{u, o\} \). If the similarity between $I$ and $T_{o}$ is higher than the similarity with the $T_{u}$, we decrease the brightness of $I$. Conversely, if the similarity between $I$ and $T_{u}$ is higher, we increase the brightness. We leverage an appropriate gamma transformation to produce the pseudo-ground truth image $I_{pgt} = 1 - (1 - I)^\gamma$, where \( \gamma \) is the factor to either enhance or reduce its brightness.

\vspace{0.5em}
\noindent\textbf{CLIP Guided Gamma Tuning.} Starting from a fixed $\gamma$, we refine the $\gamma$ per image using unpaired data and CLIP’s latent space with the well-exposed prompt $T_w$. We maximize the similarity between the pseudo-GT and $T_w$, updating $\gamma$ by the gradient of this similarity until convergence. This label-free procedure is dataset-agnostic and transferable to other exposure-related tasks.

\vspace{-8pt}
\subsection{Semantic-Prompt Consistency Loss}\label{sec:loss}
\vspace{-4pt}
We propose a Semantic-Prompt Consistency (SPC) loss as an additional objective. SPC combines the Semantic Feature Consistency (SFC) and Image-Prompt Alignment (IPA) losses in a weighted sum as follows:
\vspace{-3pt}
\begin{equation} \mathcal{L}_{\text{SPC}} = \beta_1 \mathcal{L}_{\text{SFC}} + \beta_2 \mathcal{L}_{\text{IPA}}, \end{equation}
where \(\beta_1\) and \(\beta_2\) are balancing weights. 

\vspace{0.5em}
\noindent\textbf{Semantic Feature Consistency Loss.} Our SFC loss leverages the semantic prior of the FastSAM to ensure consistency of semantic regions between the output image and the pseudo-GT. Inspired by ECLNet’s \cite{36} treatment of exposure levels as distinct styles, we leverage semantic features and the Gram matrix to ensure consistent exposure correction across semantic regions:
\begin{equation}
\scalebox{0.9}{$
\mathcal{L}_{SFC} = \sum_{i}\left( \frac{D\left(h_f^i, h_{g}^i\right)}{D\left(h_f^i, h_{g}^i\right) + D\left(h_f^i, h_{l}^i\right)} + 
\frac{G\left(h_f^i, h_{g}^i\right)}{G\left(h_f^i, h_{g}^i\right) + G\left(h_{f}^i, h_{l}^i\right)} \right),
$}
\end{equation}
where \(D\) is the L1 distance and \(G\) the Gram-matrix difference; \(h_f^i\), \(h_g^i\), and \(h_l^i\) are semantic features of the output, pseudo-GT, and input at channel \(i\). The sum runs over all semantic channels.

\vspace{0.5em}
\noindent\textbf{Image-Prompt Alignment Loss.} We introduce IPA loss, leveraging CLIP’s visual-language priors. Specifically, we maximize the cosine similarity between each output image and the fine-tuned well-exposed prompt \(T_w\), while minimizing similarity to overexposed (\(T_o\)) and underexposed (\(T_u\)) prompts. This encourages the final output to more accurately align with well-exposed characteristics and is expressed as:
\vspace{-5pt}
\begin{equation}
\label{eq:ipa-loss}
\mathcal{L}_{\mathrm{IPA}} 
= \sum_{i \in \{u, o\}} \log\Bigl(1 + \exp\bigl(sim_i - sim_w\bigr)\Bigr)
\end{equation}
\vspace{-5pt}

\noindent\textbf{Overall Loss Function.} We additionally include an MSE term to enforce intensity fidelity and a cosine term to encourage color consistency in sRGB. The complete loss function is expressed as:
\vspace{-5pt}
\begin{equation} \label{eq:loss}
\scalebox{0.9}{$ \mathcal{L}_{\text{TOTAL}} = \lambda_1 \mathcal{L}_{\text{MSE}} + \lambda_2 \mathcal{L}_{\text{COS}} + \lambda_3 \mathcal{L}_{\text{SPC}} $} ,
\end{equation}

\noindent where $\lambda_1$, $\lambda_2$, and $\lambda_3$ are the balancing weights.

\begin{table*}[htbp]
\centering
\scriptsize
\resizebox{\textwidth}{!}{%
\begin{tabular}{l*{12}{c}}
\toprule
\multirow{3}{*}{Methods} & \multicolumn{6}{c}{MSEC \cite{29}} & \multicolumn{6}{c}{SICE \cite{sice}} \\
\cmidrule(lr){2-7} \cmidrule(lr){8-13}
                        & \multicolumn{2}{c}{Under} & \multicolumn{2}{c}{Over} & \multicolumn{2}{c}{Average} 
                        & \multicolumn{2}{c}{Under} & \multicolumn{2}{c}{Over} & \multicolumn{2}{c}{Average} \\
\cmidrule(lr){2-3} \cmidrule(lr){4-5} \cmidrule(lr){6-7} \cmidrule(lr){8-9} \cmidrule(lr){10-11} \cmidrule(lr){12-13}
                        & PSNR$\uparrow$ & SSIM$\uparrow$ & PSNR$\uparrow$ & SSIM$\uparrow$ & PSNR$\uparrow$ & SSIM$\uparrow$
                        & PSNR$\uparrow$ & SSIM$\uparrow$ & PSNR$\uparrow$ & SSIM$\uparrow$ & PSNR$\uparrow$ & SSIM$\uparrow$ \\
\midrule
ZeroDCE (CVPR20) \cite{bib9}        & 18.2030 & 0.8002 & 7.2810  & 0.5511 & 12.7420 & 0.6757 & 15.7729 & 0.5502 & 5.9362  & 0.3870 & 10.8545 & 0.4686 \\ 
RUAS (CVPR21) \cite{bib14}           & 13.8135 & 0.7693 & 5.2131  & 0.4296 & 9.5133  & 0.5995 & 16.6306 & 0.5712 & 4.5444  & 0.3330 & 10.5875 & 0.4521 \\ 
EnlightenGAN (TIP21) \cite{bib10}    & 13.9441 & 0.7567 & 8.4574  & 0.6424 & 11.2008 & 0.6996 & 17.0176 & 0.6453 & 5.9867  & 0.5148 & 11.5022 & 0.5801 \\ 
SCI (CVPR22) \cite{bib17}           & 10.9230 & 0.6979 & 5.0354  & 0.4463 & 7.9792  & 0.5721 & 17.8687 & 0.6524 & 4.4517  & 0.3416 & 11.1602 & 0.4970 \\ 
PairLIE (CVPR23) \cite{bib16}       & 12.4261 & 0.7045 & 7.3216  & 0.5748 & 9.8739  & 0.6396 & 16.6697 & 0.6254 & 6.2621  & 0.3978 & 11.4659 & 0.5116 \\ 
CLIP-LIT (ICCV23) \cite{bib22}      & 17.8663 & 0.8049 & 9.5558  & 0.6699 & 13.7111 & 0.7374 & 15.1130 & 0.6111 & 7.5403  & 0.4445 & 11.3267 & 0.5278 \\
NeRCo (ICCV23) \cite{20}            & 18.2953 & 0.7858 & 14.6369 & 0.7344 & 16.4661 & 0.7601 & 18.3092 & 0.6238 & 11.9949 & 0.5163 & 15.1521 & 0.5701 \\
PSENet (WACV2023) \cite{42}         & 19.1069 & 0.8382 & 17.6407 & 0.8320 & 18.3738 & 0.8351 & 17.5358 & 0.6490 & 12.5097 & 0.5402 & 15.0228 & 0.5946 \\
LightenDiffusion (ECCV24) \cite{lightdiff} & \second{19.7651} & \best{0.8418} & 12.5553 & 0.7582 & 16.1602 & 0.8000 & \second{19.0193} & \best{0.6757} & 9.3065  & 0.5002 & 14.1629 & 0.5879 \\
UEC (ECCV24) \cite{47}             & 18.5293 & 0.8308 & \second{19.1223} & \second{0.8424} & \second{18.8258} & \second{0.8366} & 16.8998 & 0.6416 & \second{16.5405} & \second{0.6568} & \second{16.7202} & \second{0.6492} \\
\midrule
\textbf{Ours}           & \best{20.0607}      & \second{0.8392}       & \best{19.8717}      & \best{0.8527}       & \best{19.9662}      & \best{0.8460}       & \best{19.4152}      & \second{0.6706}       & \best{18.0584}      & \best{0.7006}       & \best{18.7368}      & \best{0.6866}       \\ 
\bottomrule
\end{tabular}
}
\vspace{-4pt}
\caption{Quantitative comparisons with unsupervised methods. “Under” and “Over” denote the underexposed and overexposed subsets, and “Average” is their mean. Higher PSNR and SSIM indicate better image quality. Best results are shown in \textbf{bold} and second-best are \underline{underlined}.}

\vspace{-15pt}
\label{com_un}
\end{table*}

\vspace{-10pt}
\begin{table}[ht]
\centering
\scriptsize
\resizebox{\columnwidth}{!}{%
\begin{tabular}{lcccccc}
\toprule
\multirow{2}{*}{Methods} & \multicolumn{3}{c}{MSEC \cite{29}} & \multicolumn{3}{c}{SICE \cite{sice}} \\
\cmidrule(lr){2-4} \cmidrule(lr){5-7}
 & LPIPS ($\downarrow$) & BRISQUE ($\downarrow$) & NIMA ($\uparrow$) & LPIPS ($\downarrow$) & BRISQUE ($\downarrow$) & NIMA ($\uparrow$) \\
\midrule
ZeroDCE \cite{bib9}       & 0.4348 & 42.8249 & 4.6148 & 0.5271 & 45.2872 & 4.5431 \\
RUAS \cite{bib14}         & 0.5047 & 45.8697 & 4.4174 & \second{0.2461} & 25.8244 & 4.8999 \\
EnlightenGAN \cite{bib10} & 0.3114 & 30.1323 & 4.8953 & 0.3643 & 27.2067 & 4.8252 \\
SCI \cite{bib17}          & 0.3346 & 32.7715 & 4.8284 & 0.7537 & 56.2747 & 4.0070 \\
PairLIE \cite{bib16}      & 0.3871 & 38.7468 & 4.3494 & 0.4318 & 36.5174 & 4.3053 \\
CLIP-LIT \cite{bib22}     & 0.2843 & 32.2784 & 4.6556 & 0.3677 & 31.9127 & 4.4517 \\
NeRCo \cite{20}          & 0.3248 & 36.2093 & 4.9761 & 0.4002 & 30.9285 & 4.8740 \\
PSENet \cite{42}         & \best{0.2104} & 30.1071 & 4.9901 & 0.2807 & 30.0326 & 4.7881 \\
LightenDiff \cite{lightdiff} & 0.2357 & \second{29.5824} & \second{4.9951} & 0.3146 & \second{22.9735} & 4.4875 \\
UEC \cite{47}           & 0.2309 & 35.4327 & 4.9703 & 0.3033 & 27.7842 & \best{4.9513} \\
\midrule
\textbf{Ours}            & \second{0.2107} & \best{27.6882} & \best{5.0131} & \best{0.2117} & \best{15.8901} & \second{4.9228} \\
\bottomrule
\end{tabular}%
}
\vspace{-5pt}
\caption{Quantitative comparisons in terms of additional metrics.}
\vspace{-15pt}
\label{tab:additional}
\end{table}

\section{Experiments}
\label{sec:typestyle}
\vspace{-4pt}

\vspace{-4pt}
\subsection{Experimental Setup}\label{sec:Experimental Setup}
\vspace{-4pt}

\begin{figure}[h]
    \centering

    \begin{subfigure}[b]{0.24\linewidth}
        \includegraphics[width=\linewidth]{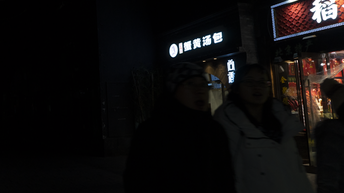}
        \caption{Input}
    \end{subfigure}
    \begin{subfigure}[b]{0.24\linewidth}
        \includegraphics[width=\linewidth]{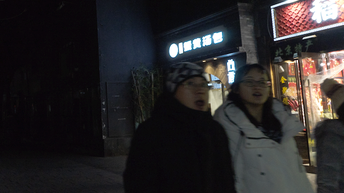}
        \caption{Output}
    \end{subfigure}
    \begin{subfigure}[b]{0.24\linewidth}
        \includegraphics[width=\linewidth]{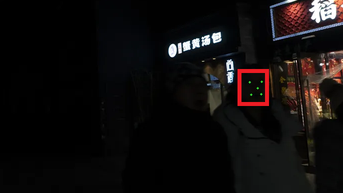}
        \caption{Input’s result}
    \end{subfigure}
    \begin{subfigure}[b]{0.24\linewidth}
        \includegraphics[width=\linewidth]{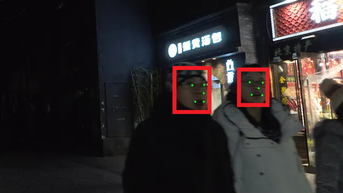}
        \caption{\scriptsize Output’s result}
    \end{subfigure}
    \vspace{-8pt}
    \caption{Face detection results on DarkFace \cite{darkface} dataset.}
    \vspace{-12pt}
    \label{fig:face}
\end{figure}

\noindent\textbf{Datasets.} We train on MSEC~\cite{29} and SICE~\cite{sice}. We use only the under-/over-exposed subsets of MSEC (2{,}830 train / 193 test). For SICE, the 2nd and last levels are treated as under/over inputs and a middle level as GT (512 train / 30 test). For prompt fine-tuning, we sample well/under/over images from SICE (152/146/161) and MSEC (213/228/205). Moreover, our correction reduces missed detections in low-light face detection(Fig.~\ref{fig:face}) on DarkFace~\cite{darkface} dataset.

\noindent\textbf{Implementation Details.} PyTorch on one NVIDIA A100. Adam ($\beta_1{=}0.9$, $\beta_2{=}0.99$), lr $1{\times}10^{-4}$, batch 8. Inputs resized to $384{\times}384$ with random horizontal/vertical flips.

\vspace{-8pt}
\subsection{Comparisons with State-of-the-art Methods}
\vspace{-4pt}

\noindent\textbf{Quantitative Comparisons.}
In the comparisons with unsupervised methods (Table~\ref{com_un}), our model outperforms nearly all compared methods, including CLIP-LIT~\cite{bib22}, PSENet~\cite{42}, and the recent UEC~\cite{47}. Specifically, we achieve substantial improvements over the previous best model, UEC, on two datasets. Then in Table~\ref{tab:additional}, we further provide additional quantitative comparisons of unsupervised methods across multiple metrics, including LPIPS~\cite{lpips}, BRISQUE~\cite{BRISQUE}, and NIMA~\cite{nima}.

\noindent\textbf{Qualitative Comparisons.} Figure~\ref{fig:comparison} presents qualitative comparisons on representative images from the MSEC and SICE datasets using state-of-the-art unsupervised methods and our approach. Existing methods often struggle under mixed lighting, over-brightening bright regions or oversaturating colors. In contrast, our method restores natural colors and preserves structures while balancing exposure across regions.

\vspace{-10pt}
\begin{table}[h!]
\centering
\small
\resizebox{\linewidth}{!}{%
\begin{tabular}{llcccccc}
\toprule
\multirow{2}{*}{} & \multirow{2}{*}{Methods} 
& \multicolumn{2}{c}{Under} 
& \multicolumn{2}{c}{Over} 
& \multicolumn{2}{c}{Average} \\
\cmidrule(lr){3-8}
& & PSNR & SSIM & PSNR & SSIM & PSNR & SSIM \\
\midrule
\multirow{2}{*}{Module} 
& w/o ASF         & 18.94 & 0.649 & 17.75 & 0.680 & 18.35 & 0.664 \\
& w/o SpatialAttn & 19.16 & 0.652 & 17.92 & 0.700 & 18.54 & 0.676 \\
\addlinespace[2pt]
\midrule
\multirow{3}{*}{Loss} 
& w/o SPC         & 19.09 & 0.643 & 17.73 & 0.667 & 18.41 & 0.655 \\
& w/o IPA         & 18.96 & 0.641 & 17.82 & 0.675 & 18.39 & 0.658 \\
& w/o COS         & 19.31 & 0.665 & 18.01 & 0.694 & 18.66 & 0.679 \\
\midrule
& \textbf{Ours} 
& \textbf{19.41} & \textbf{0.671} 
& \textbf{18.06} & \textbf{0.703} 
& \textbf{18.74} & \textbf{0.687} \\
\bottomrule
\end{tabular}
}
\vspace{-5pt}
\caption{Ablation on modules and loss components.}
\vspace{-10pt}
\label{tab:ablation_merged}
\end{table}

\vspace{-8pt}
\begin{figure}[h!]
    \centering

    \begin{subfigure}[b]{0.24\linewidth}
        \includegraphics[width=\linewidth]{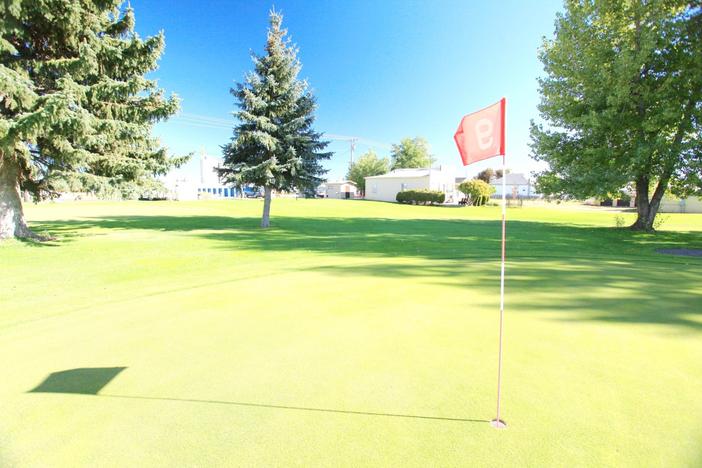}
        \caption{Input}
    \end{subfigure}
    \begin{subfigure}[b]{0.24\linewidth}
        \includegraphics[width=\linewidth]{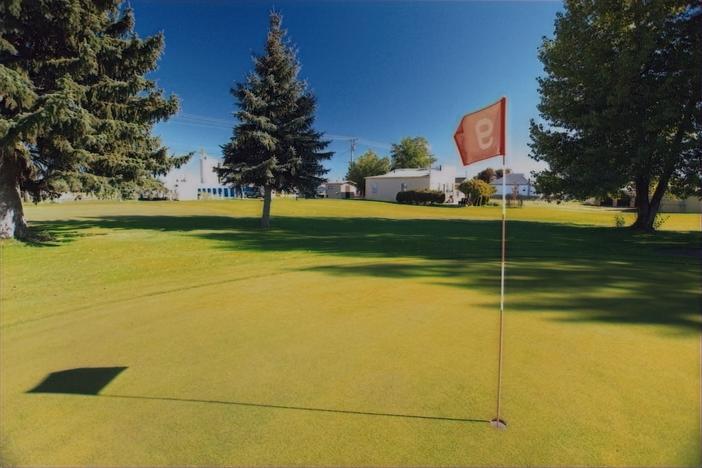}
        \caption{w/o ASF}
    \end{subfigure}
    \begin{subfigure}[b]{0.24\linewidth}
        \includegraphics[width=\linewidth]{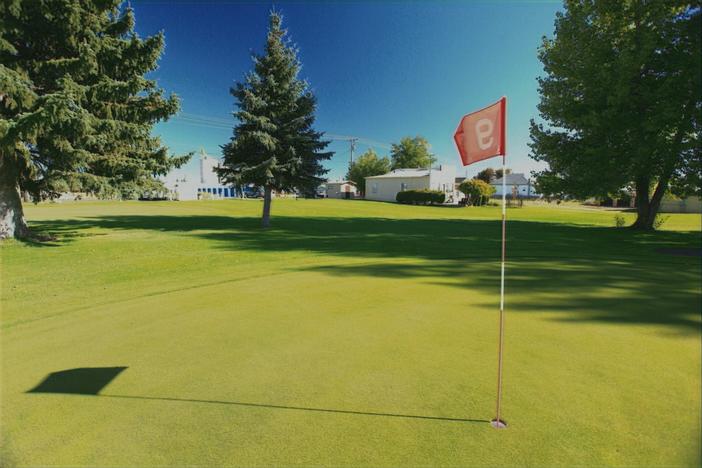}
        \caption{w/ ASF}
    \end{subfigure}
    \begin{subfigure}[b]{0.24\linewidth}
        \includegraphics[width=\linewidth]{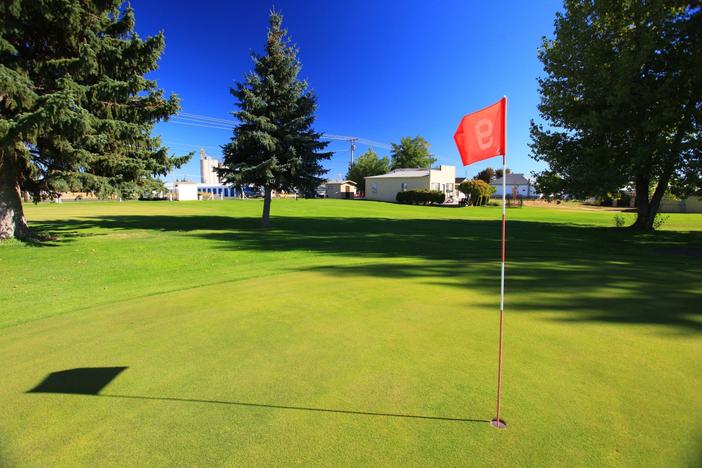}
        \caption{Reference}
    \end{subfigure}
    \vspace{-8pt}
    \caption{Effect of the ASF module.}
    \vspace{-15pt}
    \label{fig:asf_ablation}
\end{figure}

\vspace{-8pt}
\subsection{Ablation Study}
\vspace{-4pt}

We conduct comprehensive ablations on modules and losses in Tables~\ref{tab:ablation_merged} on the SICE dataset~\cite{sice}. It shows that removing the ASF module causes the largest drop in both PSNR and SSIM, highlighting the importance of semantic-guided feature integration(see Fig.~\ref{fig:asf_ablation}). Removing the SpatialAttn branch also degrades performance, confirming the necessity of spatial attention for robust correction under mixed lighting. It further dissects the loss design: discarding SPC or IPA leads to notable declines. Overall, each component contributes meaningfully, and the full model achieves the best scores.
\vspace{-10pt}

\section{Conclusion}
\label{sec:majhead}
\vspace{-8pt}
In this work, we propose an unsupervised semantic-aware framework for exposure correction that delivers superior visual quality and improved metric scores. Extensive experiments demonstrate the effectiveness and robustness of our approach. While effective, inference speed slows for very high-resolution inputs. For future work, we plan to leverage generative models for localized inpainting to recover details lost in over/underexposed regions and explore lighter backbones and pruning/distillation to cut latency and memory. Moreover, extending exposure correction to dynamic video and multi-camera settings could benefit from recent advances in camera motion understanding and large-scale camera array calibration~\cite{lin2025camerabench,you2025multi}.

\vspace{-7pt}
\begingroup
\vspace{-7pt}
\small                 %
\setlength{\bibsep}{2pt} %

\makeatletter
\renewcommand{\refname}{References\vspace*{-0.5\baselineskip}}
\makeatother

\bibliographystyle{IEEEbib}
\bibliography{strings,refs}
\endgroup

\end{document}